\documentclass[letterpaper]{article} 
\usepackage{aaai19} 
\usepackage{times}
\usepackage{helvet} 
\usepackage{courier} 
\usepackage[hyphens]{url} 
\usepackage{graphicx} 
\usepackage{amsmath,amssymb,amsfonts}
\usepackage{algorithmic}
\usepackage{textcomp}
\usepackage{xcolor}
\usepackage{tabularx}

\title{Building Second-Order Mental Models for Human-Robot Interaction}
\author {Connor Brooks$^{1}$, Daniel Szafir$^{1,2}$\\
$^{1}$Department of Computer Science, $^{2}$ATLAS Institute \\
University of Colorado Boulder\\ 
\{connor.brooks, daniel.szafir\}@colorado.edu
} 

\begin{document}

\maketitle

\begin{abstract}
The mental models that humans form of other agents---encapsulating human beliefs about agent goals, intentions, capabilities, and more---create an underlying basis for interaction. These mental models have the potential to affect both the human's decision making during the interaction and the human's subjective assessment of the interaction. In this paper, we surveyed existing methods for modeling how humans view robots, then identified a potential method for improving these estimates through inferring a human's model of a robot agent directly from their actions. Then, we conducted an online study to collect data in a grid-world environment involving humans moving an avatar past a virtual agent. Through our analysis, we demonstrated that participants' action choices leaked information about their mental models of a virtual agent. We conclude by discussing the implications of these findings and the potential for such a method to improve human-robot interactions. 
\end{abstract}

\section{Introduction}

Humans are known to form a ``Theory of Mind'' (ToM) when interacting with other people that models their knowledge of other's beliefs and intentions \cite{baker2017rational}. Research has shown that people are able to perform this inference on internal states even from low-level action information such as movement patterns \cite{bartlett2019can}. This may have implications for human-robot interactions: humans have been shown to readily anthropomorphize and form mental models about robots \cite{mutlu2016cognitive} and even behave differently around robots for which they have different mental models. For instance, humans maintain varied proxemic distances for the same robot based on its likeability \cite{mumm2011human} or its movement strategy \cite{mavrogiannis2019effects}. These results suggest that humans use some type of ToM when interacting with robots and that these models may affect human actions around robots.

%In order for robots to better interact with and understand humans around them, robots may need an element of ``self-awareness:'' a consideration for how the robot's own presence affects these people and their decision making due to people using a Theory of Mind about the robot. A robot that simply predicts the actions of nearby humans, without considering how the robot itself and how it is viewed by these humans might factor into their plans, may struggle to accurately understand human activities. For instance, a robotic car that is driving aggressively may model other drivers as extremely passive if it does not incorporate the possibility that other drivers are behaving passively to avoid the robot's own perceived unsafe driving style. Furthermore, a robot that can actively update its planning around humans based on the human's mental model of the robot could enable desirable features, such as trust calibration and appropriate explanation of its actions. 

If humans form a ToM about robots, we could enable robots to estimate how a human is affected by a robot by using \textit{second-order mental modeling} --- a robot forming a belief over a human's model of the robot. A robot or agent might acquire such a model by directly asking a human about their mental model of the agent, but this is an intrusive and potentially inefficient method of gathering information. Another option is to model the human's belief updates from some initial belief state based on their assumed observations, forming an \textit{open-loop model} of the human's beliefs. However, this assumes knowledge of both what observations the human is making and, importantly, what their starting belief about the robot was before the observations began. We hypothesize that information about a human's mental model of a robot might be gained from observing the human's actions, enabling inference of second-order mental models.

Intuitively, a human's low-level actions around another agent (such as a robot) may contain information that could be used to understand how they internally view the agent. To further illustrate this idea, consider a scenario in which people must walk past a robotic arm placing objects into a box. In this case, the interaction between the human and robot is quite brief; however, the human's mental model of the robot can still affect their choice of actions. If a person is confident that the robot will not move outside the radius of the box in the near future, this person may choose a path that narrowly misses the robot in order to preserve efficiency of their own movements. Another person that believes the robot might soon change its task and swings its end effector to some unknown new location will likely give the robot a wider berth. In these cases, the humans' differing models of the robot agent's task-based intention and future movements cause them to choose different actions despite the humans sharing the same goal.

%This scenario also illustrates an important reason why it may be beneficial to infer a human's view of a robot from their actions instead of relying only on the aforementioned open-loop modeling. An open-loop model requires assuming priors representing the human's initial beliefs about the robot before the interaction begins, then updates these priors based on the observations the human is assumed to have made. Depending on knowledge of a human's prior beliefs about the robot introduces many potential issues, as people may have different inherent biases or expertise working with the robot. Producing an accurate inference of a human's mental model of a robot from their actions could help provide insight into their current view of a robot without relying on assumptions about their initial knowledge or beliefs.

%Alternative - To connect this principle of humans using theory of mind back to the stated goal of understanding human actions around robots, let us consider three different types of human-robot interaction scenarios. In collaborative settings, a human's model of a robot teammate will inform... In cooperative or passive settings... In adversarial settings...%

If humans form mental models of agents, their actions may contain information that could be used to infer these models. The question we address in this work is whether low-level human action choices during human-agent interactions ``leak'' information \cite{mutlu2009nonverbal} that could be used in human-robot interactions to allow robots to perform Bayesian inference of a second-order ToM model. In this paper, we describe a framework for actively updating a belief of a human's model of an agent based on their actions. Then, we conduct an online study in which humans move a virtual character past a virtual ``robot'' agent and test our framework's inference of a person's model of the virtual agent from their movement choices during this interaction. We also use the data from this study to evaluate if our framework improves estimation of the human's goal. Our results show that information can be extracted from these low-level actions, although inference of mental models directly from observed actions may be noisy.

\section{Background}
This work is part of an emerging body of research focused on modeling human decision processes by extracting information from their actions. In particular, our work builds on research into human ToM and methods for learning representative models.

\subsection{Modeling Humans for Action Understanding}
Prior work has found that correctly predicting human actions through an understanding of human intent may produce safer human-robot interactions \cite{lasota2017survey} and more efficient human-robot teams \cite{kanno2003method}. In addition, \citeauthor{choudhury2019utility} \shortcite{choudhury2019utility} found that robot planning that utilizes a model of human planning may produce robot plans that generalize better than plans learned without such a model. Toward this end, past work has developed methods for predicting human intent based on low-level human motion \cite{thompson2009probabilistic,kelley2012deep,perez2015fast} as well as higher-level models of human reasoning based on modeling social forces \cite{luber2010people} or using inverse planning to explain observed actions \cite{gray2005action,baker2009action,sohrabi2016plan}.

Instead of directly inferring intent, other approaches perform inverse planning on a model of human decision making. For example, maximum entropy inverse reinforcement learning centers around finding the parameterization of a reward function that best explains a human's actions, with the assumption that the human is acting approximately optimally to maximize that reward function \cite{ziebart2008maximum}. Other frameworks for modeling human actions incorporate intention as part of a larger mental model of the human \cite{hiatt2017human}. The intention-driven dynamics model (IDDM) developed by \citeauthor{wang2013probabilistic} \shortcite{wang2013probabilistic} models human intent as a latent variable that affects the transition function of the environment. \citeauthor{trafton2013act} \shortcite{trafton2013act} create a cognitive architecture for modeling how humans process information, using this to help robots interacting with humans understand how humans will complete various tasks. \citeauthor{hiatt2011accommodating} \shortcite{hiatt2011accommodating} use such an architecture to simulate different beliefs and goals that the human could hold in order to find which set of beliefs and goals best explain actions taken by the human that were unexpected by the robot.

Mental models of humans developed to explain human actions can also model factors in the human mental state beyond goal or intent. For instance, \citeauthor{reddy2018you} \shortcite{reddy2018you} infer a human's internal dynamics model by assuming suboptimal behaviors when controlling an agent are a result of incorrect internal dynamics. \citeauthor{herman2016inverse} \shortcite{herman2016inverse} develop a gradient-based method for learning both the human's internal dynamics and reward function simultaneously. \citeauthor{tabrez2019explanation} \shortcite{tabrez2019explanation} identify issues in a human's reward function that can then be used to provide coaching to a human teammate. Individualization can be done through modeling human psychological factors, such as personality factors \cite{bera2017sociosense}, grouping people into ``types'' \cite{nikolaidis2015efficient}, or inferring people's navigation strategies \cite{godoy2016moving}.

Models of a human's mental state can also be used to estimate a human's current knowledge or beliefs \cite{breazeal2009embodied,talamadupula2014coordination,devin2016implemented,nikolaidis2017game}. Bayesian ToM \cite{baker2011bayesian} models how humans perform this inference, casting a human as an observer making inferences about an agent's beliefs and intentions. One example of this is work by \citeauthor{lee2019bayesian} \shortcite{lee2019bayesian} that estimates a child's belief about a robot's current state (listening vs not listening) during a storytelling interaction. Their approach accomplishes our goal of actively inferring a mental model, which they achieve through a filtering process that includes both a forward estimate of the person's belief after a new observation and an update based on the person's actions. However, their model is an example of a \textit{first-order} ToM-based model --- forming a model of a person's belief about factual knowledge (in this case, the robot's attentive state).

Each of these approaches use models of human reasoning to explain and predict human actions. An agent utilizing these models could be said to be practicing first-order ToM by attributing mental states to a human and modeling these states to understand human actions. However, this kind of modeling does not incorporate humans' potential views regarding agents themselves as intelligent actors about which humans may form a ToM. Our interest is in performing Bayesian inference of second-order ToM-based models from human actions around an agent, thereby learning about a human's mental model of the agent.

\subsection{Modeling Humans' Use of Theory of Mind}

Second-order mental modeling relies on the premise that humans practice first-order ToM by forming mental models of others. Modeling a human's use of first-order ToM can improve understanding of their actions, as demonstrated by \citeauthor{baker2017rational} \shortcite{baker2017rational}. Further demonstrating this benefit, \citeauthor{jara2016naive} \shortcite{jara2016naive} create a computational model of how humans reason about others' decision making based on inferring costs and rewards. This principle has been used to include a ToM in simulation agents, creating better simulations of humans in social scenarios \cite{pynadath2005psychsim,doshi2010modeling,stuhlmuller2014reasoning}. \citeauthor{rabinowitz2018machine} \shortcite{rabinowitz2018machine} create a neural network inspired by ToM that is able to build models of various types of observed agents which can be used for applications such as predicting false beliefs. Models incorporating higher-order ToM better explain human actions in competitive games than models not including ToM \cite{yoshida2008game}, and modeling how humans reason in multiplayer games can give insight into how people form social strategies such as cooperating or competing \cite{kleiman2016coordinate}. 

Moving beyond first-order ToM, \citeauthor{yang2018bayes} \shortcite{yang2018bayes} found that using higher-order ToM for policy recognition in competitive games provides an advantage over an opponent using a lower-order ToM. In line with these findings, humans tend to use first-order ToM, but can raise this level based on identifying intelligent opponents that are using higher-order ToM in competitive games \cite{doshi2010modeling,de2017negotiating}. In general, exploring beyond first-order ToM has shown that there are some benefits to humans using second-order ToM in competitive settings, but there are diminishing returns for higher orders beyond this \cite{de2014agent}.

In order to perform second-order mental modeling, some recursive structure must be used to model another agent modeling oneself (for a survey of the many ways in which agents can model one another, see \citeauthor{albrecht2018autonomous} \shortcite{albrecht2018autonomous}). \citeauthor{gmytrasiewicz1995rigorous} \shortcite{gmytrasiewicz1995rigorous} established a framework for recursive agent models, which were expanded on using techniques for belief updating in these models \cite{gmytrasiewicz1998bayesian}. \citeauthor{gmytrasiewicz2005framework} \shortcite{gmytrasiewicz2005framework} further developed these ideas into Interactive Partially Observable Markov Decision Processes (I-POMDPs), which provide a principled framework for finding policies while considering recursive beliefs about other agents. Following work by \citeauthor{doshi2012modeling} \shortcite{doshi2012modeling} that models human recursive reasoning in a game through an I-POMDP, we leverage the I-POMDP framework to model a human's internal planning around an agent.

An existing approach for building second-order mental models is what we have described above as open-loop mental modeling. \citeauthor{baker2014modeling} \shortcite{baker2014modeling} use Bayesian ToM to simulate a human inferring an agent's goal(s). \citeauthor{huang2017enabling} \shortcite{huang2017enabling} model humans as using Bayesian IRL to learn about robot objectives, and use this model to plan for actions that will communicate to the human as desired. Similarly, \citeauthor{mavrogiannis2018social} \shortcite{mavrogiannis2018social} anticipate human modeling of robot motion in order to create legible navigation strategies. \citeauthor{peltola2018modelling} \shortcite{peltola2018modelling} use an assumed second-order mental model of a recommender system for interpreting feedback from a user, while \citeauthor{gray2014manipulating} \shortcite{gray2014manipulating} instead use a second-order mental model in an adversarial manner by planning to fool a human competitor. Such research demonstrates that second-order mental modeling has the potential to improve performance of autonomous systems in interactive settings involving humans. These prior approaches are open-loop in nature: they use an assumed human mental model of the robot and update this model according to the human's observations of the robot's actions, but do not incorporate feedback into their model from the human's actions. 

The primary idea of our contribution is to create and test the feasibility of a framework for performing Bayesian inference of our estimate of the human's mental model of an agent based on the human's actions. We do this by framing the interaction between a human and an agent as an I-POMDP, then inferring the human's belief about their interactive state. This allows us to estimate the human's belief over the space of agent mental models without having to learn the human's entire transition function --- i.e., we can model the human's uncertainty about the agent while still assuming the human knows the effect of actions on the environment. Furthermore, we demonstrate an approach for using this technique to update our estimate of the human's goal or reward function. Thus, our contribution produces both an estimate of a human's internal model of another agent and an estimate of human goals while around this other agent. We anticipate that this approach could be used to jointly provide a systematic way to update beliefs about how a human views a robot during an interaction and improve understanding of human actions around robots.

\section{Inferring Human Beliefs and Goals from Actions}
In order to approach this problem, we set up an Interactive-POMDP to model human planning during an interaction. We apply tools developed for inverse reinforcement learning to this framework in order to update our beliefs about the human's mental model of an agent from observed human actions. Additionally, we derive a way to use this framework for estimating human goals while taking into account possible mental models of the agent that may be affecting their actions.

\subsection{Interactive-POMDPs for Model Inference}
The I-POMDP framework \cite{gmytrasiewicz2005framework} modifies a POMDP to include other agents by creating the notion of an \textit{interactive state}. Interactive states contain both environmental state and the states (or models) of other agents. We use a simplified version of an I-POMDP that assumes the environment is fully observed by both agents and that models of other agents are \textit{subintentional} \- models which are not recursive and instead directly map states to actions.

Formally, let an I-POMDP representing the human's perspective be given by the tuple $\left( IS, A, T, R\right)$, where $IS = (S \times M)$ is a set of interactive states, representing both environment state $s \in S$ and the agent model type attributed to the other agent $m_{R} \in M$. $A$ is a set of actions $a \in A$ available to both agents, $T$ is the transition function that updates the interactive state given human action $a_{H}$ and agent action $a_{R}$ via $T(s' | s, a_{H}, a_{R})$, and $R$ is the human's reward function. 

%Since we are using subintentional models, the environment state $s$ and agent model $m_{R}$ completely determine the probability of robot actions. This framework is presented for a human and one other agent (the robot), but the framework could be extended to accommodate multiple other agents.

Given this framework, our desired contribution is to infer part of the human's belief about their interactive state. Specifically, we want to infer $m_{R}$ given an environment state-action trajectory $\xi^T = [(s^1,a_{H}^{1}),...,(s^T,a_{H}^{T})]$. That is, after observing $\xi^T$ during a human-agent interaction, we want to determine the probability $p(m_{R} | \xi^T)$ that a given mental model is influencing the human's decision making.

\subsection{Inferring Human Mental Models of Other Agents}

In order to perform inference from this trajectory, we first model the human's planning through the state and state-action value functions of our I-POMDP. We model these value functions for an interactive state $\langle s,m_{R}\rangle$:
\begin{equation}
\mbox{\fontsize{10.0}{10.0}\selectfont\(
V(\langle s,m_{R}\rangle ) = \max_{a_{H} \in A}Q(\langle s,m_{R}\rangle , a_{H})
\)}
\end{equation}
\begin{equation}
\mbox{\fontsize{10.0}{10.0}\selectfont\(
\begin{split}
Q(\langle s,m_{R}\rangle , a_{H}) =  \sum_{a_{R} \in A}p(a_{R}|\langle s,m_{R}\rangle )\Bigl[R(s,a_{H},a_{R}) \\
 + \gamma \sum_{s' \in S} T(s' | s, a_{H}, a_{R})V(\langle s',m_{R}\rangle )\Bigr]
\end{split}
\)}
\end{equation}

Following work in inverse reinforcement learning on estimating parameters of the Q-function from actions \cite{ramachandran2007bayesian}, we model the human's action selection as exponentially weighted toward actions with higher expected reward:

\begin{equation} \label{action_prob}
\mbox{\fontsize{10.0}{10.0}\selectfont\(
p(a_{H}|\langle s,m_{R}\rangle ) = \frac{1}{Z} e^{Q(\langle s,m_{R}\rangle , a_{H})}
\)}
\end{equation}
where $Z$ is a normalizing factor.

We use this estimate along with an assumption of stationarity of belief (i.e., the human is not actively updating their mental model of the agent or reward function during this interaction) to obtain the likelihood of the observed trajectory:

\begin{equation} \label{iterative_update}
\mbox{\fontsize{10.0}{10.0}\selectfont\(
p(\xi | m_{R}) = \Pi_{(s^{i},a_{H}^{i})}p(a_{H}^{i}|\langle s^{i},m_{R}\rangle ) 
\)}
\end{equation}

% Connor -
% Originally I wrote the above equation to include a normalizing factor for the trajectory: \frac{1}{Z_{\xi}} .
%Looking at it now, I think this is unneeded, but may want to verify this again before camera ready

In order to do inference on the human's mental model type, we reverse this probability:
\begin{equation} \label{model_prob}
\mbox{\fontsize{10.0}{10.0}\selectfont\(
p(m_{R} | \xi) = \frac{p(\xi | m_{R}) p(m_{R})}{p(\xi)}
\)}
\end{equation}

Since we use a uniform prior on all model types, for a given $\xi$, the most likely model can be found by finding the model that maximizes the probability of the observed trajectory.
%\begin{equation}
%p(m_{R} | \xi) \propto p(\xi | m_{R})
%\end{equation}
Using the iterative update implied by Eq. \ref{iterative_update}, we can update our estimate of the human's mental model type as we observe new actions during an interaction.

\subsection{Joint Inference of Human Mental Model and Reward}
The prior section assumes full knowledge of the human's reward function in calculating the state-action value function. However, in many cases, we may not know either the human's mental model of the agent or the human's goal. In these cases, we can jointly reason over both possible mental models and possible goals.

To do this, let us assume we have some set of possible goals, $\phi \in \Phi$, that are used to parameterize the human's reward function: $R_{\phi}(s, a_{H}, a_{R})$. We denote the state-action value function for a particular goal as $Q_{\phi}(\langle s,m_{R}\rangle, a_{H})$. Rewriting Eq. \ref{action_prob} and Eq. \ref{iterative_update} with $Q_{\phi}(\langle s,m_{R}\rangle, a_{H})$ produces the probability of a trajectory jointly conditioned on model and goal: 

\begin{equation} \label{iterative_update_2}
\mbox{\fontsize{10.0}{10.0}\selectfont\(
p(\xi | m_{R}, \phi) = \frac{1}{Z_{\xi}}\Pi_{(s^{i},a_{H}^{i})}e^{Q_{\phi}(\langle s^{i},m_{R}\rangle , a_{H}^{i})}
\)}
\end{equation}

This gives us the joint probability of model type and goal:

\begin{equation} \label{joint_prob}
\mbox{\fontsize{10.0}{10.0}\selectfont\(
p(m_{R}, \phi | \xi) = \frac{p(\xi | m_{R}, \phi)p(m_{R}, \phi)}{p(\xi)}
\)}
\end{equation}

This joint probability can be used to find either the probability of model type or goal by marginalizing out the other variable. In our experiment, we marginalize across a discrete set of model types to find the probability of a goal from an observed trajectory:

\begin{equation} \label{goal_prob}
\mbox{\fontsize{10.0}{10.0}\selectfont\(
p(\phi | \xi) = \sum_{m_{R} \in M} p(m_{R}, \phi | \xi)
\)}
\end{equation}

\section{Evaluation}
\begin{figure}
\centering
\includegraphics[width=0.65\columnwidth]{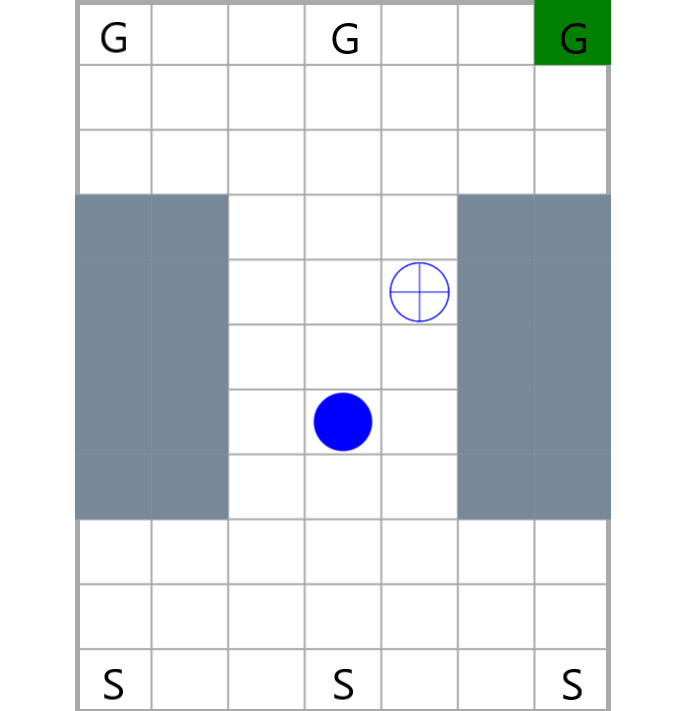}
\caption{The grid-world game played by participants. The participant's character is the filled-in circle, while the robot agent is represented by the crossed circle. The current goal is represented as a green, filled-in square. For illustrative purposes, all three possible goal locations are labeled with ``G'' labels, and the three possible participant starting locations are denoted with ``S'' labels. } \label{GameLayout}
\end{figure}

We conducted a $4 \times 1$ within-subjects study to collect data on how humans incorporate mental models of autonomous agents into planning. We recruited $102$ participants from Mechanical Turk for this study. Turkers were required to have at least a $90\%$ approval rate, and we limited the participants to U.S. IP addresses to limit language translation issues with our experiment instructions. During the experiment, participants control an avatar that moves through a grid-world in order to reach a designated goal. This grid-world contains a second agent as well: a virtual ``robot'' agent with which the participant is required to avoid collisions. This grid-world and the interface used by participants is shown in Fig. \ref{GameLayout}. This task is meant to simulate a simple, passive human-robot interaction in which a human needs to move past a robot in order to continue along the human's intended path. 

\subsection{Experimental Procedure}
Participants complete 5 trials per condition. For each trial, participants are given one of three goals on the top row of the grid world. The positions of these three possible goals requires the participant to move their avatar past the agent, which starts randomly in the 3x3 section directly in the middle of the grid-world. Participant and agent movements are done in a turn-based manner at discrete time steps. There are 4 different strategies for agent movement, creating 4 conditions:

\smallskip

\noindent \textit{Stationary Agent:} In this baseline condition, the virtual  agent simply stays still for the duration of the trial.

\noindent \textit{Random Agent:} The virtual agent chooses actions purely randomly.

\noindent \textit{Fixed-Goal Agent:} The virtual agent moves toward its goal, which is at the participant's starting location.

\noindent \textit{Chasing Agent:} The virtual agent chases the participant by choosing actions at each state that move it closer to the human agent. Equivalently, the agent moves toward a goal at the participant's current location.

We design our experiment to ensure that participants have an accurate mental model of the agent before each trial begins so that we can test our hypotheses using this known mental model as the ground-truth. To accomplish this, before each condition's trials begins, participants are informed of the strategy that the agent will be using to select actions. After this, participants complete a practice trial in which they move through the grid-world, including moving past the appropriate agent used in the next set of trials. Once participants end this practice trial, they answer a single question to ensure they correctly understand the agent's movement strategy. This question consists of displaying a sample game state and asking the participant in which the direction the agent would move next based on the agent type for the upcoming trials. If participants incorrectly answer this question, they are taken back to the description of the ``robot'' agent model; this loop continues until participants answer correctly. The purpose of this procedure is to ensure the participants are internally using the correct mental model of the agent for each condition. Additionally, the practice trial serves to demonstrate the agent behavior to the participant so that they do not act according to surprise or reflexive behavior during the recorded trials. 

If the participant collides with the agent during a trial, the trial is reset. Participants complete 5 successful trials with each condition before moving to the next (only trajectories from successful trials were used in our analysis). Finally, the ordering of conditions is randomized in order to mitigate biasing from ordering effects.

\section{Hypotheses}
We hypothesize that our framework for Bayesian inference of second-order mental models will enable improved estimation of both a human's view of an agent and their goal:

\smallskip

\noindent \textit{Hypothesis I:} Given a known goal, human actions will leak information that can be utilized by second-order mental modeling to infer the mental model the human holds of the agent.

\noindent \textit{Hypothesis II:} Estimation of a human's goal via marginalizing the joint probability of goal and agent mental model will lead to improved estimation compared with directly estimating the goal while assuming the human treats the agent merely as an obstacle.

\section{Analysis}
The purpose of our study setup is to record trajectories in which we know the human's mental model of the agent is correct due to the explanation and practice period before each trial. This allows us to test our framework's ability to infer this ground-truth mental model directly from the human action sequences. 

First, we investigate our primary question of whether our framework can infer the human's mental model of the agent from their actions by evaluating the accuracy of what our framework judges to be the most likely model based on an observed trajectory with a known goal. To do this, we set uniform priors over the four agent models used for our four conditions, then update these probabilities based on a trajectory. We calculate accuracy based on whether or not the agent model assigned the highest probability at the end of the trajectory matches the correct model for that trial.

Second, we compare the probability assigned to the correct goal over time using our method for joint estimation of agent model and goal compared to estimation of the goal while assuming the human views the agent as an obstacle (which is equivalent to using the stationary agent model at all times). We use a discrete goal space consisting of the three goals which we used in data collection: the top-left corner, top-middle, and top-right corner of the grid-world. After each action, we update and record the probability assigned to the correct goal by each method. The probabilities are presented by the percent of the trajectory observed rounded to $5\%$ intervals, averaged across all trials.

\subsection{Implementation Modifications for Turn-Based Games}
In our experiment, the human and agent take turns making movements, with each able to see the results of the other's turn before choosing their own action. In order to accommodate this type of turn-based interaction into our framework, we expand the human's state-action value functions to include the expected next turn of the agent. Essentially, this allows the Q-function to represent a full turn (both human and agent move) including both the given human action and an expectation over possible agent actions. For more compact notation, we represent the value function for a agent turn as $RV$, producing the following equations:

\begin{equation}
\mbox{\fontsize{10.0}{10.0}\selectfont\(
\begin{split}
Q_{\phi}(\langle s,m_{R}\rangle , a_{H}) = R_{\phi}(s, a_{H}, NULL) \\ 
+ \gamma \sum_{s' \in S} T(s' | s, a_{H}, NULL)RV_{\phi} (\langle s',m_{R}\rangle)
\end{split}
\)}
\end{equation}

where

\begin{equation}
\mbox{\fontsize{10.0}{10.0}\selectfont\(
\begin{split}
RV_{\phi}(\langle s,m_{R}\rangle) = \sum_{a_{R} \in A} p(a_{R} | \langle s,m_{R}\rangle) \Bigl[R_{\phi}(s, NULL, a_{R}) \\
+ \gamma \sum_{s' \in S} T(s' | s, NULL, a_{R})V_{\phi}(\langle s',m_{R}\rangle)\Bigr]
\end{split}
\)}
\end{equation}

\subsection{Tuning the Reward Function}
In order to produce the state-action value function, we need to estimate the reward assigned to state-action pairs by participants. To do this, we create a simple reward function with three parameters: reward for reaching the goal $G$, cost per action $C_{a}$, and discount factor $\gamma$. We set aside data from $10$ participants for tuning these parameters (only the data from the other $92$ participants was used for testing), then perform a grid search to find the parameters that maximize the probabilities of the observed trajectories for this set of participants. This produced the following values: $G = 30$, $C_{a} = -3$, $\gamma = 0.95$.

\section{Results}

\subsection{Inference of the Participant's Robot Model from Actions}
The results of the Top-1 and Top-2 accuracy on the trajectories from the $92$ test participants are presented in Table \ref{AccuracyTable}. An improvement in classification accuracy over the accuracy due to chance demonstrates that the actions from participants in our experiment do leak information about their mental model of the agent. 

\begin{figure}
\centering
\includegraphics[width=\columnwidth]{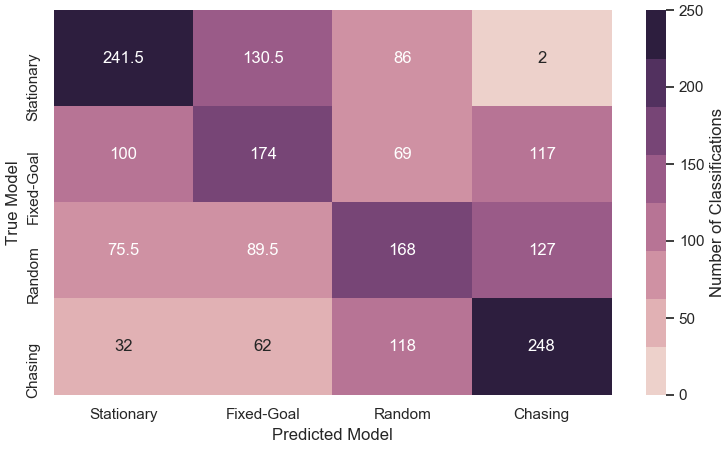}
\caption{Confusion matrix of model classification from observed trajectories.} \label{ConfusionMatrix}
\end{figure}

Fig. \ref{ConfusionMatrix} visualizes the confusion matrix of the top classification for each trial (for classifications in which there is a tie, the classification is split between all tied classes). As the confusion matrix shows, for each true model type, the correct model type is the most common classification by our method. Furthermore, some pairs of models are more frequently misclassified than other pairs --- we discuss this result in more detail below.

\begin{table}
\small
\caption{Accuracies of model classification from observed trajectories.} \label{AccuracyTable}
\begin{tabularx}{\columnwidth}{l c c }
 \multicolumn{3}{c}{Model Classification Accuracy} \\
 \hline
  & Top-1 Accuracy & Top-2 Accuracy\\
 \hline
 I-POMDP State Inference   & 45.2\%  & 78.5\% \\
 Chance &   25.0\%  & 50.0\% \\
 \hline
\end{tabularx}
\end{table}

\subsection{Estimation of the Participant's Goal}
\begin{figure}
\centering
\includegraphics[width=\columnwidth]{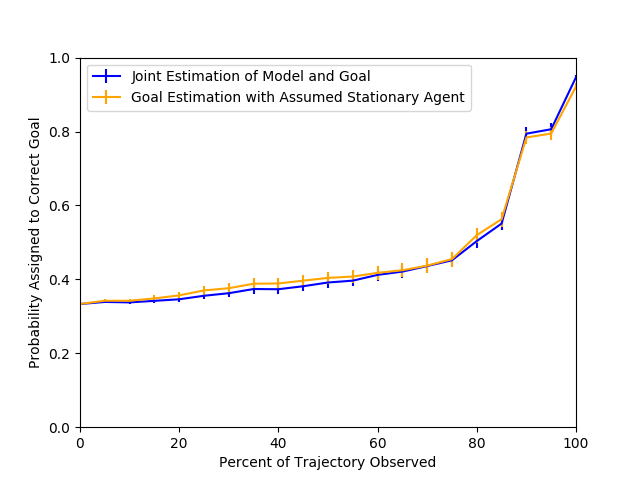}
\caption{The probability assigned to the true goal by the percent of the trajectory observed.} \label{GoalProbabilities}
\end{figure}

Fig. \ref{GoalProbabilities} visualizes the probability assigned to the correct goal over percent of the trajectory observed, averaged across all trials. An initial uniform probability of the three possible goals is assumed, which is then updated at each timestep according to the procedure laid out previously. In both cases, the correct goal is nearly always inferred with high confidence as the human approaches the goal. Our intent in visualizing these probabilities over time through the trajectories is to examine if joint inference of the model and goal has an effect on this inference as compared to assuming a fixed agent location at each action step (Hypothesis 2). We found that the two methods performed similarly, with the assumed stationary agent slightly outperforming joint inference in initial action observations, while the joint inference method assigns slightly higher probabilities to the correct goal as more actions have been observed.

\section{Discussion}
There are a few key takeaways from the results of our analysis. First, low-level actions do provide information about a human's mental model of a local agent, even without considering the human's observations. This is shown through our I-POMDP state inference method outperforming the accuracy due to chance at classification of the participant's ground-truth mental model of the agent. However, our method is not robust in this scenario; the classifier only selects the correct model approximately half of the time. Nevertheless, the results demonstrate that there is potential to learn these second-order mental models from human actions around agents such as robots, even during brief, passive interactions. Furthermore, these second-order mental models are learned using a model of actions as independent choices (as opposed to considering trajectories which could implicitly encode the behavior of the agent). This shows that information about the participants' view of the agent is contained directly in some subset of individual action selections, even without taking into account the agent's actual movement or behavior. Thus, we found support for Hypothesis 1, although more work is needed to improve classification accuracy.

Second, an interesting pattern is shown by the confusion matrix from the classification (Fig. \ref{ConfusionMatrix}). The confusion matrix demonstrates that the classification errors are predominantly between the pairs of Stationary and Fixed-Goal models, and Random and Chasing models. Intuitively, these pairs of models represent the agent which are ``safe'' to pass directly next to (the Stationary and Fixed-Goal models) due to either not moving or having a pre-determined path that the participant can easily avoid, and the agents which are ``unsafe'' to pass directly adjacent to (the Random and Chasing models) due to the potential for collision from uncertain or hostile agent models. This clustering demonstrates the potential for actions to leak information not only about specific models that the human holds about an agent, but perhaps also higher-level information about more abstract mental models that might encompass multiple specific models.

Finally, the benefit to goal inference from our method is unclear. While our joint model and goal inference method produces marignally (but not significantly) higher final probabilities assigned to the correct goal compared with treating the agent as a fixed object in the view of the human while modeling action probabilities, the difference is quite small and overall the two methods perform similarly. Consequently, we do not find support for Hypothesis 2. Further testing on more complex interaction scenarios is required to determine whether or not there is a benefit to using this method for goal inference.

\section{Future Work}
Through this paper, we have laid out a formulation for what we hope will be a useful way to improve understanding of how humans view robots during an interaction. Our experiment demonstrates that there is indeed information embedded in low-level human actions that can be used for second-order mental modeling. However, further work is needed to apply these ideas to real-world human-robot interactions.

First, in this paper, we have proposed a framework for performing Bayesian inference of second-order mental models and investigated the feasibility of doing so from low-level human actions. In order to establish this approach clearly, we have shown this on a simple task: a fully-observable grid-world with discrete state, action, reward, and model spaces. Relaxing these assumptions in future work would create a more powerful tool for understanding human actions without prior knowledge of possible human mental states.

Second, while we describe and evaluate our framework for Bayesian inference of second-order mental models in isolation in order to study its plausibility, an implementation in a more complex task ought to combine this framework with existing methods for open-loop modeling that take into consideration a human's observations during the interaction. Thus, this framework could serve to close the loop in an existing open-loop modeling framework. This combined framework could also relax one of the assumptions made in our formulation: that the human's belief about the agent's model type is stationary. Such a framework would first model forward propagation of belief based on the human's observations, then update this prior by taking into account the human's action according to our method developed in this paper. In this way, our method and existing open-loop modeling could be used together to utilize both human observations and human actions in updating the belief about the human's mental model without assuming either a certain initial model or a stationary model.

Finally, we evaluate our framework in this paper using a human-agent interaction scenario with greatly simplified dynamics and interaction possibilities. While we have developed this framework with human-robot interactions in mind, this initial study only directly demonstrates leaking of information about mental models in an online human-agent interaction.  We look forward to conducting further studies with humans and physically-embodied robots pariticipating in passive, teaming, and adversarial interactions in order to further examine whether this leakage occurs and is useful in these scenarios. 

\section{Conclusion}
If humans use a Theory of Mind about robots during interactions, second-order mental modeling might provide information that could be used by robots to be aware of their effect on the people around them. We have shown that human action choices around agents in a grid-world environment do leak information about a human's view of such an agent. In doing so, we have provided a justification for closing the loop in open-loop mental models of humans in order to take into account this information as human actions around robots are observed.

A robot planning with an awareness of its effect on humans provides many possible benefits. For instance, such a robot could identify errors with a human's view of the robot that might lead to sub-optimal interactions, then communicate to fix these errors in order to improve the remainder of the interaction. Actively assessing a human's view of a robot provides a path for deepening and improving understanding between humans and robots during human-robot interactions.  

\fontsize{9.0pt}{10.0pt}
\selectfont
\bibliography{references}
\bibliographystyle{aaai}
\end{document}